\documentclass[twocolumn,letterpaper]{IEEEAerospaceCLS}  
\usepackage[font=small]{caption}
\usepackage[]{amsmath,amssymb,dsfont,color,graphicx,subcaption,mathtools,nccmath,bm,enumitem,caption}
\usepackage{url}

\begin{document}
\title{ReachBot: A Small Robot for Large Mobile Manipulation Tasks}

\author{%
Stephanie Schneider\\ 
Dept. of Aero. \& Astronautics\\
Stanford University\\
496 Lomita Mall\\
Stanford, CA 94305\\
schneids@stanford.edu
\and 
Andrew Bylard\\ 
Dept. of Aero. \& Astronautics\\
Stanford University\\
496 Lomita Mall\\
Stanford, CA 94305\\
bylard@stanford.edu
\and
Tony G. Chen\\ 
Dept. of Mech. Engineering\\
Stanford University\\
440 Escondido Mall\\
Stanford, CA 94305\\
agchen@stanford.edu
\and
Preston Wang\\ 
Dept. of Aero. \& Astronautics\\
Stanford University\\
496 Lomita Mall\\
Stanford, CA 94305\\
wpreston@stanford.edu
\and 
Mark Cutkosky\\ 
Dept. of Mech. Engineering\\
Stanford University\\
440 Escondido Mall\\
Stanford, CA 94305\\
cutkosky@stanford.edu
\and
Marco Pavone\\ 
Dept. of Aero. \& Astronautics\\
Stanford University\\
496 Lomita Mall\\
Stanford, CA 94305\\
pavone@stanford.edu
\thanks{\footnotesize 978-1-6654-3760-8/22/$\$31.00$ \copyright2022 IEEE}           
}

\maketitle

\thispagestyle{plain}
\pagestyle{plain}

\maketitle

\thispagestyle{plain}
\pagestyle{plain}

\begin{abstract}
Robots are widely deployed in space environments because of their versatility and robustness. However, adverse gravity conditions and challenging terrain geometry expose the limitations of traditional robot designs, which are often forced to sacrifice one of mobility or manipulation capabilities to attain the other.
Prospective climbing operations in these environments reveals a need for small, compact robots capable of versatile mobility \textit{and} manipulation. 
We propose a novel robotic concept called ReachBot that fills this need by combining two existing technologies: extendable booms and mobile manipulation. ReachBot leverages the reach and tensile strength of extendable booms to achieve an outsized reachable workspace and wrench capability.
Through their lightweight, compactable structure, these booms also reduce mass and complexity compared to traditional rigid-link articulated-arm designs.
Using these advantages, ReachBot excels in mobile manipulation missions in low gravity or that require climbing, particularly when anchor points are sparse.
After introducing the ReachBot concept, we discuss modeling approaches and strategies for increasing stability and robustness. We then develop a 2D analytical model for ReachBot's dynamics inspired by grasp models for dexterous manipulators. Next, we introduce a waypoint-tracking controller for a planar ReachBot in microgravity. Our simulation results demonstrate the controller's robustness to disturbances and modeling error.
Finally, we briefly discuss next steps that build on these initially promising results to realize the full potential of ReachBot.
\end{abstract}

\tableofcontents    

\section{Introduction} \label{sec:intro}
The intersection of mobility and manipulation 
stands out as a key technology need for the future of space exploration. Mobile manipulators have proven to be essential on missions to space, such as the Mars rover~\cite{RobinsonCollinsEtAl2013}, but new robotic paradigms are needed to extend these capabilities to more challenging and unpredictable environments.
In particular, there is a growing interest in operations that require robots capable of mobile manipulation under a variety of gravity regimes and terrain geometries that are not conducive to traditional robotic traversal. For example, in microgravity or the very low gravity characteristic of small bodies such as comets and asteroids, gravity is insufficient to maintain contact with a surface without anchoring. The same applies in environments that require climbing under non-negligible gravity, where robots must be anchored to avoid sliding or falling off steep, vertical, or overhanging surfaces. In these environments, anchoring is a key component of long-term stability, safety, and mobility.

\begin{figure}[tp!]
    \centering
    \includegraphics[width=0.45\textwidth]{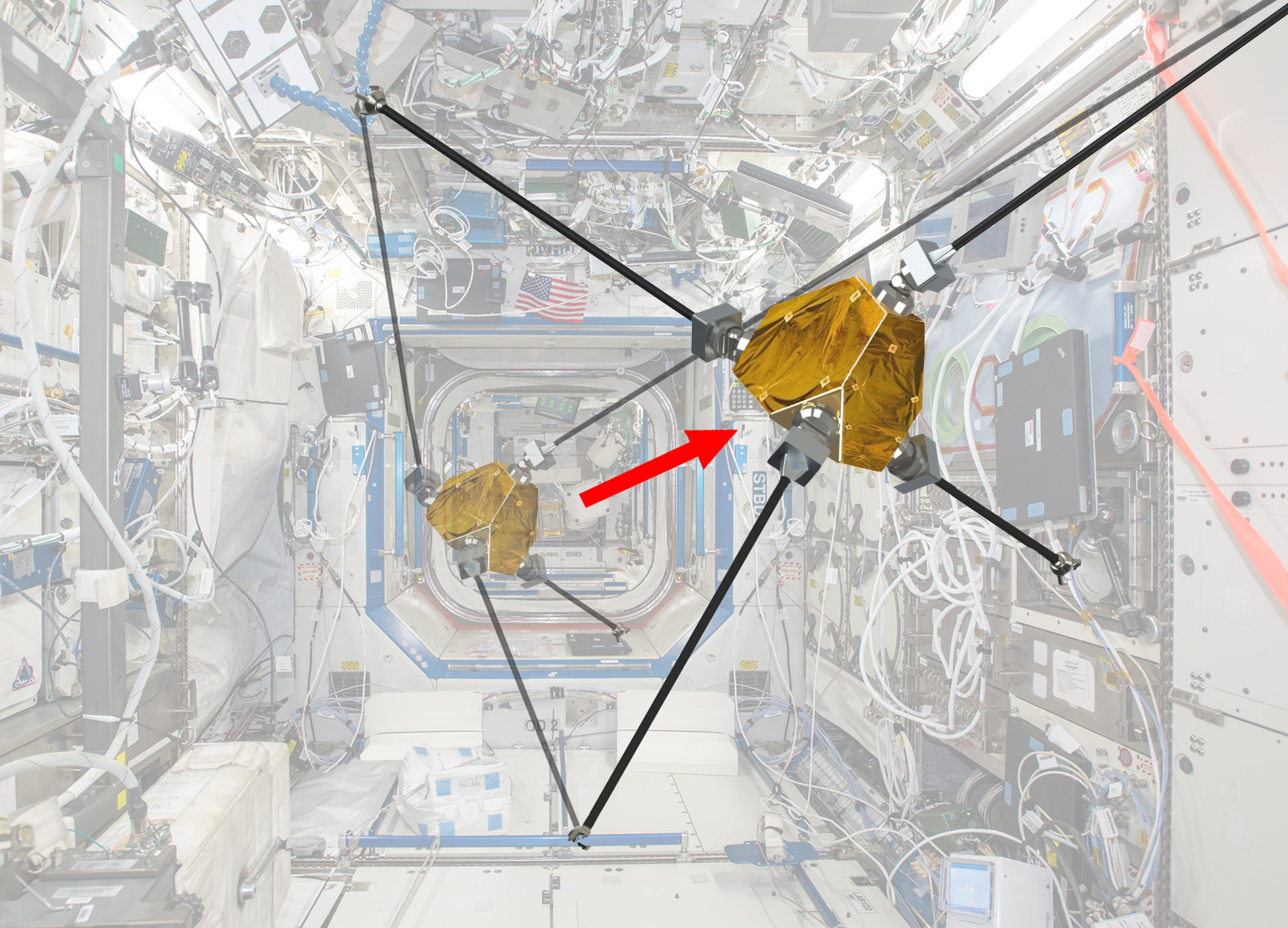}
    \caption{\label{fig:iss} A ReachBot robot traverses a space station module. Unlike small robots that are limited by short reach or large robots limited by mass and bulk, ReachBot is particularly well-suited to on-orbit tasks. The lack of gravity-induced bending forces allows the booms to reach extreme lengths while having small diameters and thin walls, leading to compactable booms and spool mechanisms that have a small footprint and low mass.}
    \vspace{-10pt}
\end{figure}

In addition to mobility, anchoring in these environments provides a secure foundation from which a robot can perform forceful manipulation. Without the cooperation of gravity, neither friction nor the weight of the robot can be relied on to provide stabilization during high-wrench manipulation tasks. Thus being able to reach and leverage a large variety of anchor points becomes crucial for manipulation in these environments.

Relevant missions in microgravity include on-orbit construction and both intra- and extra-vehicular space station servicing. For example, the Lunar Gateway mission proposes a space station that must sustain long periods of uncrewed operation, requiring robots to perform all maintenance and logistics tasks~\cite{CoderreEdwardsEtAl2019}. 
Tasks on station such as pulling velcroed objects from a wall can require high manipulation forces, but become trivial for robots that can simultaneously leverage anchor points on the floor, ceiling, and opposing walls, as shown in Figure~\ref{fig:iss}.
Additionally, the interior surface of a space station is often prized real estate, ideally devoted to science-related and mission-critical equipment. 
This prioritization of wall space motivates long-reach robot paradigms that require only a sparse set of anchor points for all required tasks.
 
Similarly, in the very low gravity of small bodies such as comets and asteroids, anchored mobility and manipulation are essential for targeted surface exploration and precision in-situ inspection. The highly irregular geometries and loosely attached material permitted by such low gravity may also lead to secure anchor points being sparse and with haphazard placement, calling for versatile long-reach robots.

Relevant missions that involve climbing under gravity include exploring the caverns and lava tubes on the moon and Mars. These have been identified as areas of great geological interest, providing a preserved record of ancient material in stratified layers that could give insights into the history of the solar system~\cite{LeveilleDatta2010,Cushing2017}. Additionally, these sheltered sites could contain key astrobiological targets~\cite{Boston2010} and are potential sites for future human habitation.
To access scientific targets in different layered strata, a robot would need to navigate steep, vertical, or overhanging surfaces on cavern walls and ceilings where reliable anchor points may be sparse.
Additionally, such a robot must have high-wrench capability to drill into rock, both to take in-situ measurements and potentially collect samples for return to Earth. In situations where targets are in tight spaces or are far from viable anchor points, a long-reach robot would also be able to deliver lightweight science instruments on the ends of its arms.

These application domains illustrate the need for space robots that combine sparse-anchored mobility with high-wrench manipulation, but there is a key technology gap in existing solutions.
Small robots are typically restricted to a small reach and limited wrench capability. Consequently, they may have difficulty reaching anchor points needed for mobility or performing manipulation tasks that require a large wrench force. 
Conversely, large robots, particularly rigid-link articulated-arm robots, are hampered by high mass and complexity, which scale poorly with increased reach.
Rather than sacrificing maneuverability or reach, in this paper we propose a novel robotic platform that combines the portability of small mobile robots with the long reach and high-force capability of large robots.

Our concept, called ReachBot, introduces the novel combination of extendable booms and mobile manipulation. Figure~\ref{fig:iss} shows ReachBot navigating around a space station. Extendable booms have traditionally been used in space for deployable structures, such as solar sails~\cite{Fernandez2017}, telescope membranes~\cite{FootdaleMurphey2014}, or solar panels~\cite{SpenceWhiteEtAl2018}. These booms are ideal for space applications because their composite triangular or slit-tube structure keeps them lightweight and compact when rolled up, while still being strong in tension and capable of extending many meters, particularly in reduced gravity.
ReachBot repurposes a set of extendable booms as prismatic joints, using them as arms both for mobility and to perform high-wrench manipulation.
The innovative application of this technology to a new context enables ReachBot to achieve versatile mobility and forceful interaction in challenging environments.

\textit{Statement of Contributions:} This paper addresses the need for mobile manipulation in previously inaccessible space environments.
The main contributions of this work are threefold.
(1) We introduce ReachBot, the first robot concept that uses rollable extendable booms to support mobile manipulation.
(2) We develop a novel dynamics model for ReachBot inspired by work in the field of dexterous manipulation.
(3) We propose strategies for increasing ReachBot's robustness to disturbances, for example the impulse cause by grasp failure. We perform a preliminary robustness analysis on the design, and develop a controller for ReachBot that we validate in simulation on a planar ReachBot in microgravity.

\textit{Paper Organization:} The rest of the paper is organized as follows. In Section~\ref{sec:related}, we discuss existing robotic solutions that pursue mobility and manipulation in challenging space environments. In Section~\ref{sec:concept}, we motivate ReachBot's design, and introduce a control strategy that ensures the reliability of extendable booms in the presence of disturbance forces. Then in Section~\ref{sec:model}, we develop a dynamical model for ReachBot inspired by work in dexterous manipulation.
Section~\ref{sec:controller} presents a computed-torque controller that enables ReachBot to follow a desired series of waypoints. Following model and controller development,
Section~\ref{sec:simulations} demonstrates a simulation in which ReachBot tracks a series of waypoints, proving that a set of extendable booms is a viable paradigm for controlled mobility.

\section{Related Work} \label{sec:related}
A number of other robot solutions have been built or proposed for the environments and missions listed above.
In microgravity, such as on a space station, small assistive free-flyers such as Astrobee~\cite{BualatSmithEtAl2018}, Int-Ball~\cite{MitaniGotoEtAl2019}, and CIMON~\cite{KarraschSchmidEtAl2018} achieve precise mobility, but their weak thrusters cannot apply high loads or support maneuvering with large, high-powered attached manipulators. This limits both their anchored reach and forceful manipulation capability, making many key station servicing tasks infeasible. Additionally, free-flying robots will always have inherent risk over robots that always remain anchored, and this risk profile grows with increased robot mass and thruster strength.


Unlike free-flyers, large on-orbit articulated-arm robots such as Robonaut~\cite{AmbroseAldridgeEtAl2000,DiftlerMehlingEtAl2011} offer much more manipulation capability and secure mobility in microgravity, but this comes at a high cost of mass and mechanical and electrical complexity--issues which have been an ongoing obstacle for Robonaut. The crawling robot arm on the Tiangong space station~\cite{XueLiuEtAl2021}, as well as the Dragonfly arm concept~\cite{Ridinger2017}, are compelling paradigms for long-reach on-orbit mobile manipulation, but they require large launch volume and are limited by fixed-length links.


In very low-gravity environments such as on asteroids and comets, the main mobility paradigm for surface rovers is hopping robots. This had been explored in projects such as Hedgehog~\cite{HockmanPavone2017,Hockman2018} and was used with great success by MINERVA-II on the asteroid Ryugu~\cite{TsudaSaikiEtAl2020}. However, while hopping is extremely energy-efficient for long-distance traversal, and has been the subject of motion planning and traversability studies, these rovers do not offer a solution for precision mobility or forceful manipulation~\cite{YoshimitsuKubotaEtAl2003}, both of which are important for targeted scientific exploration. Their hops also impart impulses on the environment that could disturb natural formations which would otherwise be targets of careful analysis.

For climbing under gravity, LEMUR \cite{Parness2017} represents the current state of the art, using rigid-link articulated arms and microspine grippers to climb on vertical and overhanging rocky surfaces. 
However, due to the relatively small reachable workspace of its articulated arms, LEMUR is impeded in environments with sparse anchor points.
This limitation may be incompatible with environments such as underground caverns on Mars and the moon where the terrain geometry and surface material properties are difficult to predict, but may contain an abundance of gaps, loose boulders, and smooth crumbling surfaces.
Additionally, LEMUR's limited reach would prevent it from extending arm-mounted science instruments from robot body toward targets far from secure anchor points or deep into tight spaces where interesting astrobiological material is most likely to reside.
Alternatively, the Axel robot provides a wheeled rapelling design that does not rely on any anchor points.
This makes it a compelling rover for exploring craters, cliffs, and collapsed skylights of lava tubes. However, its reliance on wheel makes it unable to access overhanging surface or move reliable along a rocky floor~\cite{MatthewsNesnas2012}. Thus, it would not be able to perform complete wall stratigraphy of some cliffs and most caverns, including lava tube interiors not directly underneath collapsed roof sections.

To overcome the shortcomings of these existing designs, ReachBot must incorporate three key features. First, ReachBot must have a large reachable workspace while anchored. Second, it must be able to apply significant forces and moments, together known as wrenches. Finally, its design has to accomplish these first two objectives while limiting mass, complexity, and retracted size. In the following sections, we explain how ReachBot's unique design and dynamical modeling successfully achieve these requirements.

\section{ReachBot Concept} \label{sec:concept}
In Section~\ref{sec:concept}A, we motivate ReachBot's design by introducing the numerous advantages of using extendable booms as robot arms, particularly expanding its reach and wrench capabilities without drastically increasing mass and unextended size. In Section~\ref{sec:concept}B, we outline the challenges of this novel design and then discuss how our control strategy addresses these challenges.

\subsection{A. ReachBot Design}

\begin{figure}[tp!]
    \centering
      \includegraphics[width=0.23\textwidth]{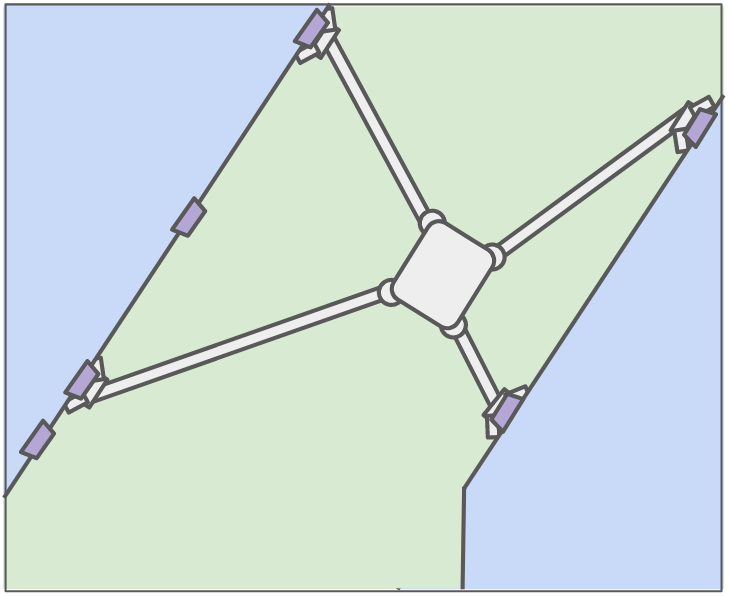}
      \includegraphics[width=0.23\textwidth]{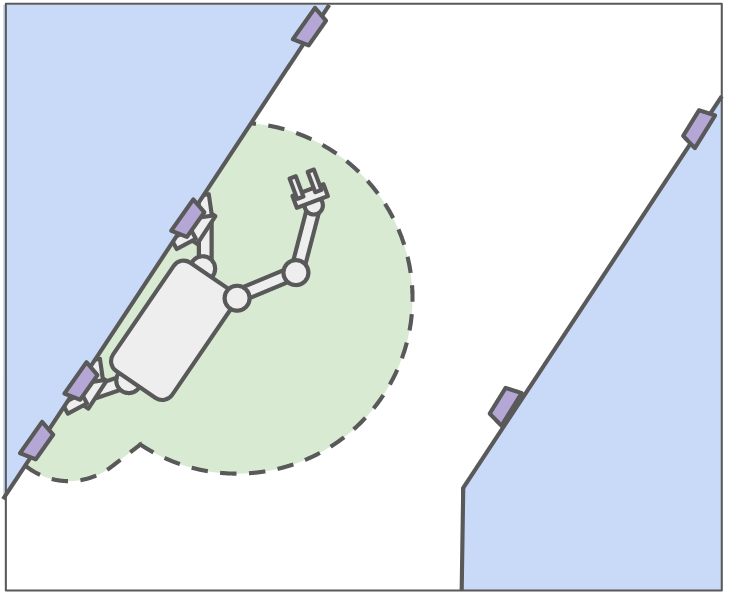}
    \caption{\label{fig:overhang} ReachBot (left) using its long reach to traverse terrain having sparse anchor points--shown in purple--that would inhibit an articulated arm robot (right). Each robot's reachable workspace is shown in green.}
    \vspace{-5pt}
\end{figure}

In contrast to traditional rigid-link robots, ReachBot provides a large reachable workspace through extendable booms. 
As shown in Figure~\ref{fig:overhang}, ReachBot's reachable space is larger than that of a comparably-sized rigid-linkdw articulated-arm robot, increasing the number of accessible anchor points.
Access to more anchor points grants ReachBot flexibility to choose between different anchoring options. 
This flexibility enables more secure grasp configurations, faster crawling speed, and less frequent re-planning, allowing ReachBot to move across terrain that might not be passable for other robots. 

In addition to enhancing mobility, ReachBot's large reachable workspace provides advantages for manipulation. 
ReachBot uses its booms as tensile members to apply large wrenches on the environment, a strategy similarly exploited by small robots that anchor themselves and pull objects using tethers~\cite{EstradaMintchevEtAl2018}. However, unlike tethers or cables, 
ReachBot's booms can be controllably extended, eliminating the need for the robot to move its body to each anchor point.

\begin{figure}[bp!]
    \centering
    \includegraphics[width=0.45\textwidth]{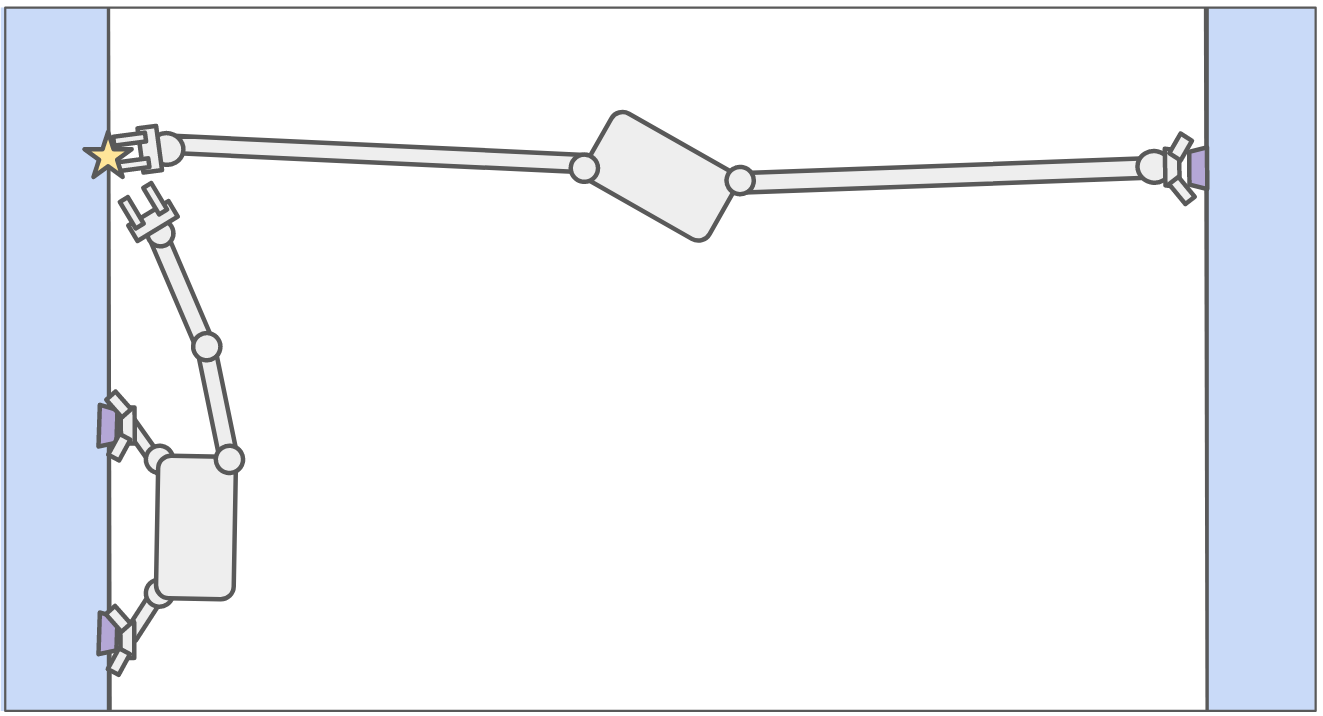}
    \caption{Alternative manipulation strategies to pull target (yellow star) away from the wall. The lower, articulated-arm robot must anchor close to the target, then apply a large motor torque to lift its outstretched arm. ReachBot, shown above, completes the manipulation task with minimal lever arm and therefore minimal motor torque. It uses an anchor on the opposite wall to apply a large manipulation force while keeping both booms under tension.}
        \label{fig:manipulation}
\end{figure}

Figure~\ref{fig:manipulation} shows a scenario in which ReachBot is able to access a target that would be challenging for an articulated-arm robot. In particular, ReachBot requires fewer anchor points and can reach the target from a much greater distance.
Additionally, ReachBot performs a high-wrench manipulation task while using less torque than alternative strategies. Instead of depending on a large, powerful motor to apply torque and bending moment across a long lever arm, ReachBot uses an anchor on an opposite surface to apply a force normal to the target surface while keeping both booms under tension.

Another key quality of ReachBot is that it achieves these advantages within a lightweight and compact design. 
Due to the lightweight single-element uniform structure of the extendable booms, ReachBot's size, mass, and complexity do not scale significantly with increased reach, unlike comparable designs that rely on higher power consumption to actuate increasingly large rigid-link arms~\cite{Parness2017}.
These features provide several advantages for space robots, including reliability, reduced launch cost, and lower energy usage. 

\subsection{B. Modelling and Control Challenges}
In this section, we describe challenges that arise from ReachBot's design and discuss how we can use intelligent control strategies to overcome these challenges.
Thin structural members like extended booms often have material properties that make them susceptible to buckling or bending failure.
However, we can overcome typical limitations by exploiting force closure and the booms' tensile strength.

We exploit ReachBot's design by leveraging existing work in force closure for dexterous manipulators. 
In the field of multi-fingered dexterous manipulation, a robotic hand pushes on an object from multiple directions, manipulating it by applying forces at the contact points between the fingertips and the object.
ReachBot's mobility is analogous to grasping with unisense contact forces~\cite{BicchiKumar2000,HanTrinkleEtAl2000}: instead of having fingers that push, we have booms that pull.
Many control strategies have been developed that enable a dexterous manipulator to follow a pose trajectory while maintaining force closure~\cite{LiHsuEtAl1989,NagaiYoshikawa1993}. 
Drawing parallels to this existing work, we can control ReachBot to achieve force closure by exploiting gravity together with the concavity of a space station or cave by establishing contact points in multiple directions.

\begin{figure}[bp!]
    \subfloat[With pretension]{%
      \includegraphics[width=0.2\textwidth]{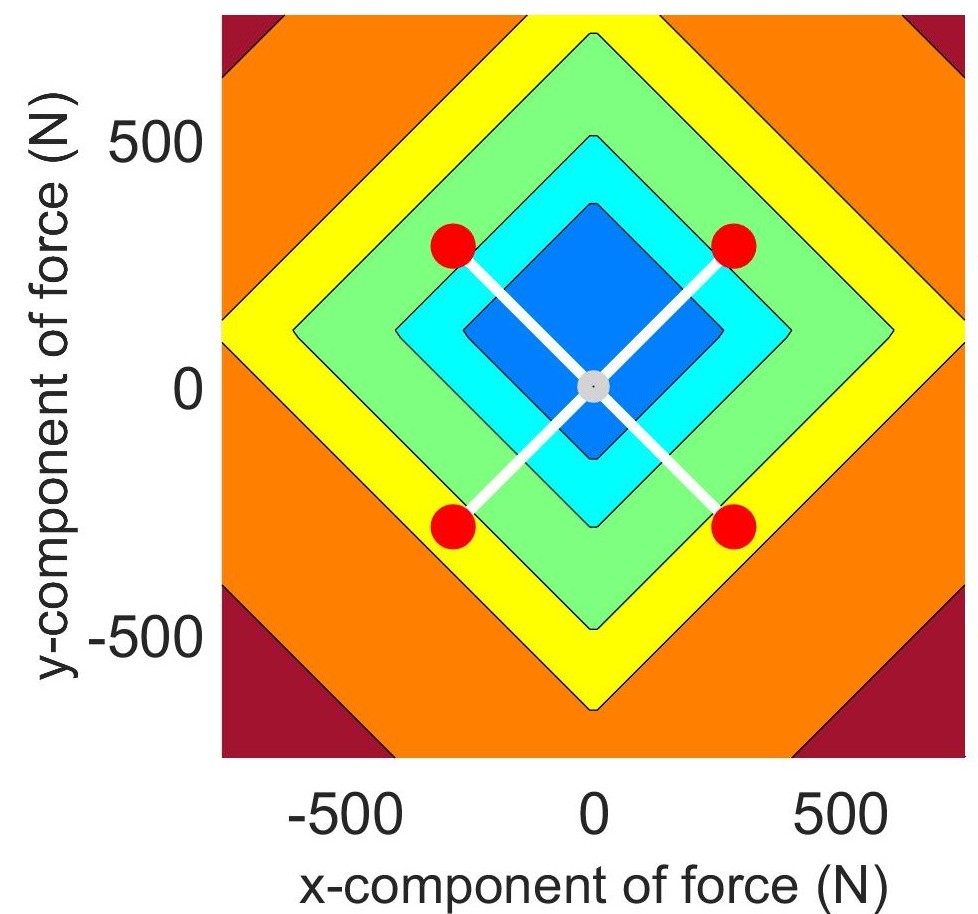}
      \label{fig:pret}}
\hspace{\fill}
   \subfloat[Without pretension]{%
      \includegraphics[width=0.24\textwidth]{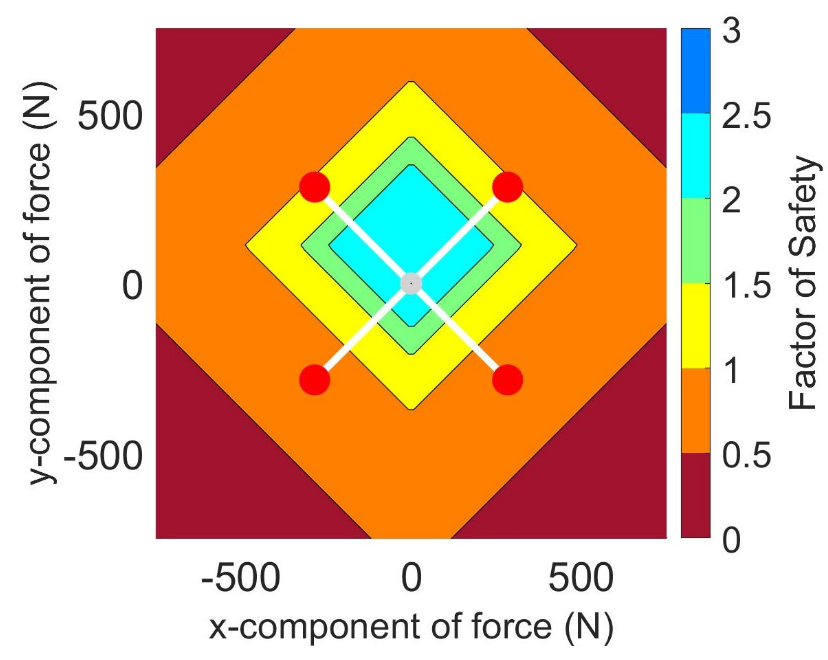}
      \label{fig:no_pret}}\\
\caption{\label{fig:pretension}
The contours show ReachBot's minimum factor of safety while resisting different disturbances, where failure is expected for any factor less than $1$ (orange and red). 
(a) With uniform pretension of $100N$ in all four booms, ReachBot has a larger factor of safety in the whole force space
(b) Without pretension, ReachBot has consistently lower factors of safety, making failure more likely.}
\end{figure}

By employing a force closure configuration,
ReachBot exploits boom tensile strength to overcome typical design shortcomings of extendable booms.
Specifically, the booms' thin walls make them susceptible to lateral disturbances and bending moments when extended. Disturbances can be caused, for example, by an impulse due to a sudden grasp failure.
Figure~\ref{fig:pretension} demonstrates how ReachBot applies uniform pretension in its booms to reduce the likelihood of failure from disturbing forces. 
The figure illustrates ReachBot's factor of safety while resisting different disturbances in two scenarios: with and without applying pretension to the booms. Failure is expected for any factor less than $1$ (shown in orange and red). In both cases, ReachBot's body configuration is stationary, and the $x$ and $y$ axes represent the $x$ and $y$ components of applied force, respectively. 
In addition to the applied force, ReachBot resists a constant downward (-$y$-direction) force of Lunar gravity. The contours in Figure~\ref{fig:pretension} show the minimum factor of safety across all booms while resisting the total force, considering both buckling and yield failure.
By leveraging the high tensile strength of the booms~\cite{Fernandez2017}, ReachBot increases its robustness to disturbances.
Figure~\ref{fig:pretension}(a) demonstrates a larger factor of safety across the whole force space when ReachBot applies a uniform pretension of $100N$ to all four booms. Figure~\ref{fig:pretension}(b) shows the same scenario without pretension, corresponding to lower factors of safety and therefore less ability to resist disturbances.
ReachBot avoids significant bending forces by leveraging the booms' tensile strength to support its structure, a strategy that has been successfully demonstrated by cable-driven parallel robots~\cite{CarricatoMerlet2012}.

In Section~\ref{sec:model}, we develop a dynamical model influenced by a proven approach for modeling dexterous manipulators. In Section~\ref{sec:controller}, we exploit a multi-fingered hand controller~\cite{LiHsuEtAl1989} to develop a controller that enables ReachBot to follow a desired series of waypoints.

\section{ReachBot Model and Dynamics} \label{sec:model}
Here we develop a model for ReachBot's dynamics and kinematics. First, in Section~\ref{sec:model}A, we introduce and justify a series of assumptions that simplify the model, including a 2D planar adaptation of ReachBot that will be used throughout the paper. 
We then break down ReachBot's motion into two alternating modes, mirroring a successful climbing paradigm for articulated-arm robots~\cite{Parness2017}. 
In Section~\ref{sec:model}B, we consider the ``body movement'' stage, where all four of ReachBot's end-effectors remain anchored to the surface while the body moves in space.
Then, in Section~\ref{sec:model}C, we consider the ``end-effector movement'' stage where ReachBot detaches an end-effector and moves it to a new anchor point. By alternating these stages of motion, ReachBot can reposition its body and arms either to navigate within the environment or to ready itself for a manipulation task.

\begin{figure}[tp]
    \centering
    \includegraphics[width=0.48\textwidth]{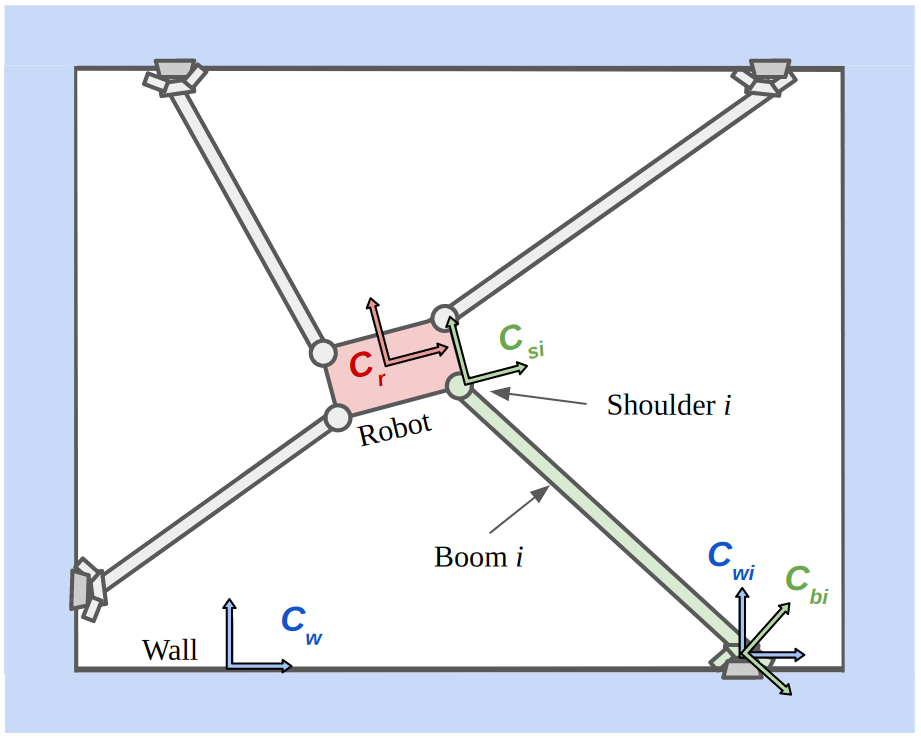}
    \caption{Coordinate frames for ReachBot in 2D. The frame $C_w$ is fixed to the inertial wall frame, while $C_r$, $C_s$, and $C_b$ correspond to the robot, shoulder joints, and booms, respectively.}
    \label{fig:ref_frames}
    \vspace{-5pt}
\end{figure}

The result of this dynamical analysis is a set of equations that determine ReachBot's motion. 
These equations of motion define the relationship between applied joint torques (forces for prismatic joints) and resulting movement of the robot. For the duration of this paper, we will refer to coordinate frames as defined in Figure~\ref{fig:ref_frames}. 
The frame $C_r$ is fixed to the robot at its center of mass. The stationary frame $C_w$ is fixed to the wall. The local wall frame $C_{wi}$ is fixed relative to $C_w$ with the origin at the point of contact with boom $i$, where the $z$-axis coincides with the wall's outward pointing normal. The shoulder frame $C_{si}$ is fixed relative to $C_r$ with its origin at the the base of boom $i$. The boom frame $C_{bi}$ is fixed to boom $i$ with its origin at the point of contact with the wall.

\subsection{A. Modeling Assumptions}
We make a number of assumptions to simplify the modeling process.
First, we assume we can ignore the dynamics of the booms and only consider the dynamics of the robot body and end-effectors. To support this assumption, we make two assertions: 
(1) the mass of the boom is negligible compared to that of the robot body and end-effector, which is consistent with our preliminary prototype, and
(2) we can treat the boom as a rigid body. The latter is consistent with an analysis we performed whereby we compared pose estimates for rigid and flexible models of a fully extended boom with a $1$kg end-effector mass and found less than a $1$\% deviation between the two models. By ignoring the boom dynamics, we simplify ReachBot's dynamical model while maintaining sufficient accuracy for a feasibility analysis.

\begin{figure}[bp]
    \centering
    \subfloat[Wall contact]{%
        \includegraphics[width=0.55\columnwidth]{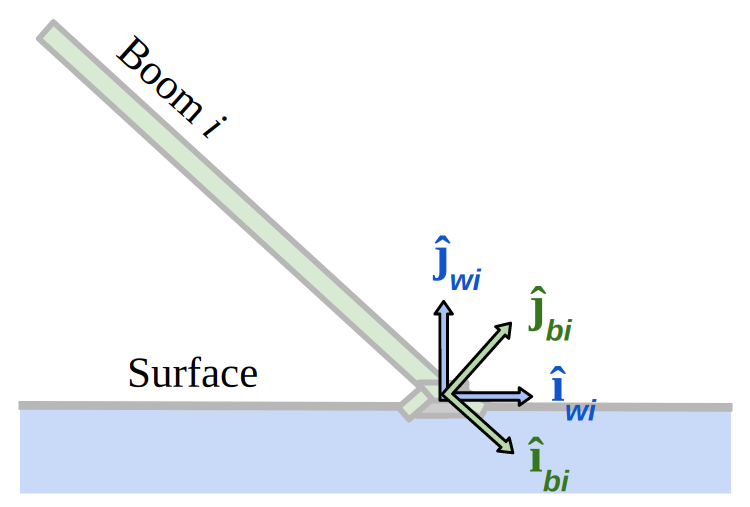}
    }
    \hspace{\fill}
    \subfloat[Embedded friction cone]{%
    \includegraphics[width=0.4\columnwidth]{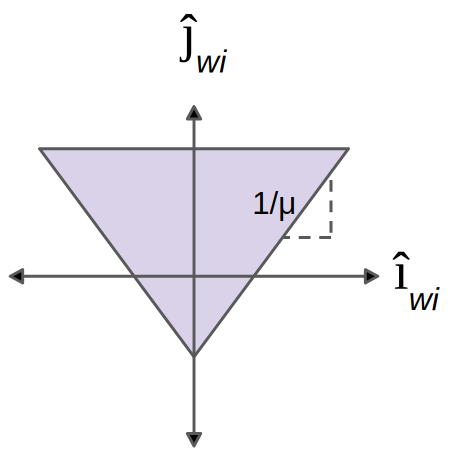}
    }
    \caption{(a) shows the contact coordinate system; boom $i$ pulls mainly in the $-\hat{i}_{bi}$ direction with normal force in the $\hat{j}_{wi}$ direction. (b) shows the embedded cone model for grip force constraints; $\mu$ is the coefficient of static friction. Any force vector in the purple region can be sustained by the gripper.}
    \label{fig:surface-grip}
\end{figure}

Second, we assume the interaction between the end-effector and the wall can be modeled as a point contact having ``embedded cone" friction~\cite{SotoParnessEtAl2008}, which is an idealization of the JKR friction/adhesion model~\cite{JohnsonKendallEtAl1971}. This friction model, illustrated in Figure~\ref{fig:surface-grip}, sustains moderate pulling forces in the tensile direction while remaining attached to the wall. 

Third, we assume there is a spring-loaded ball joint (a pin joint in the planar case) connecting the end of the boom to the gripper mount. The ball or pin joint ensures negligible moment at the contact while the spring keeps the joint from rotating freely, returning the end-effector to its equilibrium position when un-anchored.

Fourth, we assume the given trajectory can be executed with a series of stable, manipulable configurations. A configuration is said to be stable if, for every possible body wrench, there exists a choice of joint torques to balance it (up to joint torque limits). A configuration is manipulable if, for every body velocity direction, there is a choice of joint velocities that achieves the motion without breaking contact with the wall~\cite{LiHsuEtAl1989}. 
Because ReachBot remains within the convex hull of its end-effectors throughout the trajectory, and has redundantly actuated controls, there must exist a trajectory where the configurations are stable and manipulable at all times during operation. 

For simplicity, we use a 2D planar adaptation of ReachBot as a first step to validate the concept.
This planar ReachBot adaptation includes a robot body with four 
shoulder mechanisms, each having two controllable joints: (1) a prismatic joint composed of a motor that extends and retracts the rollable boom and (2) a revolute joint that pivots the boom in the plane. Each boom assembly is capped with an end-effector.
Eight independent control inputs (extension and rotation for each boom) give ReachBot full 3-DoF authority.

\subsection{B. Body Movement Model}
In this section, we derive a dynamical model for ReachBot's body when all four end-effectors are anchored to the wall.
Specifically, we develop a set of equations that represent different constraints on ReachBot's motion.
Leveraging the parallels between ReachBot and dexterous manipulators, we borrow the method from Li, et al.~\cite{LiHsuEtAl1989} to characterize the relationships between joint torques and robot velocity via the common intermediary: contact constraints.

First, we consider the dynamical constraints of the system and present equations of motion for ReachBot during its body movement stage.
Second, we constrain the end-effectors to remain in contact with the wall, resulting in both velocity and wrench constraints at the points of contact. 
Lastly, we apply the principle of virtual work to ensure each contact point stays in static equilibrium. 

\subsubsection{Dynamical constraints}
ReachBot's dynamics are given by the Newton-Euler equation as
\begin{equation}
    \label{eq:dynamics_eqn}
    \begin{bmatrix} M & 0 \\ 0 & I \end{bmatrix} \begin{bmatrix} \dot{v}_{r,w} \\ \dot{\omega}_{r,w} \end{bmatrix} +
    \begin{bmatrix} \omega_{r,w} \times v_{r,w} \\ \omega_{r,w} \times I \omega_{r,w} \end{bmatrix} = 
    \begin{bmatrix} f_r \\ \tau_r \end{bmatrix},
\end{equation}
where $M$ is the mass of the robot body multiplied by the identity matrix in ${\mathbb R}^{3 \times 3}$, $I$ is the inertia matrix with respect to the robot coordinates $C_r$, and where $v_{r,w}$ and $\omega_{r,w}$ are the linear and rotational velocity, respectively, between coordinate frames $C_r$ and $C_w$. Additionally, $[f_r^\top, \tau_r^\top]^\top$ is the resultant wrench on the robot defined in the $C_r$ frame. The resultant wrench directly causes motion, as opposed to the internal wrench, which is absorbed by the robot's mechanical structure.

\subsubsection{Zero relative motion at contact}
\begin{figure}
    \centering
    \includegraphics[width=0.3\textwidth]{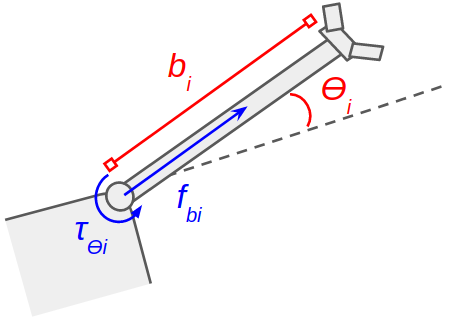}
    \caption{Model for a single boom $i$ with length $b_i$ and angle $\theta_i$ with respect to the robot frame. Controls for each boom include a prismatic joint force $f_{bi}$ and a rotational joint torque $\tau_{\theta i}$.}
    \label{fig:boom_movement_model}
    \vspace{-5pt}
\end{figure}

To ensure contact is maintained during manipulation, we derive constraint equations for both relative velocities and wrenches between the robot and the contact points.
Recall that each boom has two joints: a prismatic joint that extends and retracts the boom and a rotational joint that rotates the boom in-plane about the shoulder.
Consider the joint position vector for boom $i$, 
\begin{equation}
q_i = \begin{bmatrix} b_i \\ \theta_i \end{bmatrix},    
\end{equation}
where $b_i$ is the length of boom $i$ and $\theta_i$ is the angle of boom $i$, as shown in red in Figure~\ref{fig:boom_movement_model}. The vector $q_i$ defines the position of the boom coordinate frame $C_{bi}$ with respect to the shoulder frame $C_{si}$. The relative velocity of the shoulder frame with respect to the boom frame is related to the joint velocity vector $\dot{q}_i$ by the boom Jacobian $J_i$, defined such that
\begin{equation}
    \label{eq:boom_jacobian}
    \begin{bmatrix} v_{si,bi} \\ \omega_{si,bi} \end{bmatrix} = 
    J_i(q_i) \dot{q}_i.
\end{equation}
We model the connection between the boom and the end-effector as a pin joint~\cite{KerrRoth1986}, allowing it to rotate freely in the plane and apply forces in all directions. With the pin joint model, we can represent contact constraints between the boom and the wall using the basis matrix
\begin{equation}
    \label{eq:basis}
    B^\top_i = \begin{bmatrix}
    1 & 0 & 0 & 0 & 0 & 0 \\
    0 & 1 & 0 & 0 & 0 & 0 \\
    0 & 0 & 1 & 0 & 0 & 0 \\
    0 & 0 & 0 & 1 & 0 & 0 \\
    0 & 0 & 0 & 0 & 1 & 0
    \end{bmatrix},
\end{equation}
which is defined for boom $i$ in the $C_{wi}$ frame. 
Then, we relate the velocity of the robot in the world frame, $[v_{r,w}^\top, \omega_{r,w}^\top]^\top$, 
to the joint velocity vector $\dot{q}_i$ of each boom by
\begin{equation}
    \label{eq:vel_constraint_eq_boom}
    B^\top_i \begin{bmatrix} v_{r,w} \\ \omega_{r,w} \end{bmatrix} = 
    B^\top_i J_{wi}(q_i)\dot{q}_i,
\end{equation}
where
\begin{equation}
    J_{wi} = T_{wi,r}J_i
\end{equation}
is the boom Jacobian $J_i$ expressed in the $C_{wi}$ frame, and
\begin{equation}
    \label{eq:T}
    T_{\gamma,\beta} = \begin{bmatrix}
    A^\top_{\gamma,\beta} & -A^\top_{\gamma,\beta} S(p_{\gamma,\beta}) \\ 0 & A^\top_{\gamma,\beta}
    \end{bmatrix}
\end{equation}
is the general form of the transition matrix. In \eqref{eq:T}, $p_{\gamma,\beta} \in {\mathbb R}^3$ and $A_{\gamma,\beta} \in SO(3)$ are, respectively, the position and orientation of coordinate frame $C_\gamma$ relative to coordinate frame $C_\beta$, and $S(p)$ is the skew-symmetric matrix form of the vector $p$.

For the full four-boom manipulation system, we concatenate vectors of individual booms, mirroring the derivation for a multi-fingered hand in \cite{LiHsuEtAl1989}. We define the combined joint position vector as
\begin{equation}
    q = [q^\top_1, q^\top_2, q^\top_3, q^\top_4]^\top.
\end{equation}
We then concatenate \eqref{eq:vel_constraint_eq_boom} for all four booms, yielding an expression for the full system that relates body velocity to the combined joint velocity. The resulting velocity constraint equation is given by 
\begin{equation}
    \label{eq:vel_constraint_eq_robot}
    G^\top \begin{bmatrix} v_{r,w} \\ \omega_{r,w} \end{bmatrix} = 
    J \dot{q},
\end{equation}
where
\begin{equation}
    G = \textrm{blockdiag}(B_1, B_2, B_3, B_4)
\end{equation}
is the grasp matrix. The grasp matrix is defined as the block diagonal of individual basis matrices for each boom, and
\begin{equation}
    J = G^\top \textrm{blockdiag}(J_{w1}, J_{w2}, J_{w3}, J_{w4})
\end{equation}
is the combined robot Jacobian.

In addition to velocity constraints, we must also derive constraints on contact wrenches to ensure contact is maintained throughout manipulation.
The contact wrench at boom $i$, $x_i$, is applied along the direction of basis matrix $B_i$. Each individual contact's contribution to the wrench on the robot is given by
\begin{equation}
    \label{eq:2-8}
    \begin{bmatrix} f_r \\ \tau_r  \end{bmatrix} =
    T^\top_{wi,r} B_i x_i,
\end{equation}
where $f_r$ and $\tau_r$ represent force and torque in the robot frame, respectively, as they were defined in \eqref{eq:dynamics_eqn}. In future work, we will constrain the wrench $x_i$ based on the contact model for the gripper; in this case, $x_i$ lies in the embedded cone shown in Figure~\ref{fig:surface-grip}(b). 

The combined contact wrench vector along the directions of the basis matrices is
\begin{equation}
    x = [x^\top_1, x^\top_2, x^\top_3, x^\top_4]^\top.
\end{equation}
We constrain the total wrench on the robot by concatenating wrenches for the four booms in \eqref{eq:2-8}. This combined contact wrench constraint is given by
\begin{equation}
    \label{eq:2-13}
    \begin{bmatrix} f_r \\ \tau_r  \end{bmatrix} =
    \begin{bmatrix}T^\top_{w1,r}, T^\top_{w2,r}, T^\top_{w3,r}, T^\top_{w4,r} \end{bmatrix}
    G x.
\end{equation}
This equation expresses the resultant wrench in the $C_r$ frame, but if we expressed it in the $C_w$ frame, we would obtain the familiar relation between resultant wrench and combined contact wrench expected from the dual equation to \eqref{eq:vel_constraint_eq_robot}:
\begin{equation}
    \begin{bmatrix} f_w \\ \tau_w \end{bmatrix} = G x.
\end{equation}
\subsubsection{Static equilibrium at contact}
For each boom, the joint torque vector, which is comprised of force for the prismatic joint and torque for the rotational joint, is given by
\begin{equation}
    \tau_i = \begin{bmatrix}
    f_{bi} \\ \tau_{\theta i}
    \end{bmatrix},
\end{equation}
where $f_{bi}$ is the force applied at the prismatic joint and $\tau_{\theta i}$ is the torque at the revolute joint, as shown in blue in Figure~\ref{fig:boom_movement_model}.

By the principle of virtual work, the joint torque required by boom $i$ to maintain static equilibrium for a contact wrench $x_i$ is given by
\begin{equation}
    \label{eq:2-9}
    \tau_i = J^\top_{wi} B_i x_i.
\end{equation}
We define the combined joint torque as
\begin{equation}
    \tau = [\tau^\top_1, \tau^\top_2, \tau^\top_3, \tau^\top_4]^\top
\end{equation}
such that the torque $\tau$ required to maintain static equilibrium for combined contact wrench $x$ is
\begin{equation}
    \label{eq:2-14}
    \tau = J^\top x.
\end{equation}
\subsection{C. End-Effector Movement Model}
In the end-effector movement stage, ReachBot performs a maneuver that involves detaching an end-effector, moving the corresponding boom to a new position, and reattaching the end-effector to the wall.
By tensioning the three anchored booms, we hold the robot body motionless. Thus we only need to consider the dynamics of a single boom, which we model as a massless rod with a point mass on the end.

As before, the joint position vector is defined as
\begin{equation}
    q_i = \begin{bmatrix} b_i \\ \theta_i \end{bmatrix},
\end{equation}
where $b_i$ and $\theta_i$ are the boom length and angle, as shown in Figure~\ref{fig:boom_movement_model}. With a mass of $m_i$, the equations of motion of end-effector $i$ are given by
\begin{equation}
    \ddot{q}_i = \begin{bmatrix} \ddot{b}_i \\ \ddot{\theta}_i \end{bmatrix} =
     \begin{bmatrix} \frac{f_{bi}}{m_i} + b_i\dot{\theta}_i^2 \\ \frac{\tau_{\theta i}}{b_i m_i} - 2\dot{b}_i\dot{\theta}_i \end{bmatrix}.
\end{equation}
%

\section{Control Strategies}\label{sec:controller}
In this section, we present two controllers that jointly realize ReachBot's mobility: one for the body movement stage and one for the end-effector movement stage. 
Section~\ref{sec:controller}A discusses
the details of the body movement stage, where we modify a computed-torque controller designed for in-grasp dexterous manipulation~\cite{LiHsuEtAl1989},
tailoring it to the dynamical model of ReachBot. Then, Section~\ref{sec:controller}B defines the PD controller that supports the end-effector movement stage. 
For both stages, we present controllers that follow a desired series of waypoints and leave integration of boom tension control for future work.

\subsection{A. Body Movement Control}
Here we present a control scheme that enables ReachBot to follow a desired series of waypoints while maintaining wall contact with all four end-effectors. 
First, we define an equation for the contact wrench needed to achieve a desired wrench in the robot body frame. Then, using the relation between contact wrench and joint torques developed in Section~\ref{sec:model}B, we present a joint torque control law that enables ReachBot to realize the trajectory.

The relationship between contact wrench and resultant wrench is given by \eqref{eq:2-13}. We define $H$ as
\begin{equation}
    H = \begin{bmatrix}T^\top_{w1,r}, T^\top_{w2,r}, T^\top_{w3,r}, T^\top_{w4,r} \end{bmatrix}G .
\end{equation}
From Section~\ref{sec:model}, we assume the stances at the given waypoints and at all intermediate stances are stable, indicating that $H$ is surjective~\cite{LiHsuEtAl1989}. Consequently, we can use its left inverse, $H^\dagger$, to find the contact wrench required to produce a specific body wrench. The contact wrench $x$ required to produce body wrench $[f_r^\top,  \tau_r^\top]^\top$ is shown by
\begin{equation}
    \label{eq:li_4-7}
    x = H^\dagger \begin{bmatrix} f_r \\ \tau_r \end{bmatrix}.
\end{equation}
Recall that $[f_r^\top, \tau_r^\top]^\top$ is the \textit{resultant} wrench on the body, which produces no internal wrench. To isolate the relationship between contact wrench and robot motion, we combine \eqref{eq:dynamics_eqn} and \eqref{eq:li_4-7} to yield
\begin{equation}
    \label{eq:li_4-8}
    x = H^\dagger \left( 
    \begin{bmatrix} M & 0 \\ 0 & I \end{bmatrix} \begin{bmatrix} \dot{v}_{r,w} \\ \dot{\omega}_{r,w} \end{bmatrix} +
    \begin{bmatrix} \omega_{r,w} \times Mv_{r,w} \\ \omega_{r,w} \times I \omega_{r,w} \end{bmatrix}
    \right).
\end{equation}
%

The controller design objective is to specify a vector of joint torque inputs $\tau$ to follow a desired series of waypoints given by
\begin{equation}
    (p^d_{r,w}(t), A^d_{r,w}(t)) \in SE(3),
\end{equation}
where $p^d_{r,w}$ and $A^d_{r,w}$ are the desired pose and orientation, respectively, of the robot with respect to the wall. We locally parameterize $A_{r,w} \in SO(3)$ by $\phi_{r,w}$, which represents the roll-pitch-yaw variables such that $\phi_{r,w} = [\phi_1, \phi_2, \phi_3]^\top$ is a nonsingular parameterization of $SO(3)$~\cite{LiHsuEtAl1989}. Using this parameterization, we express desired waypoints as
\begin{equation}
        (p^d_{r,w}(t), A_{r,w}(\phi^d_{r,w}(t))) \in SE(3),
\end{equation}
and the body velocity as 
\begin{equation}
    \label{eq:li_4.1-5}
    \begin{bmatrix} v_{r,w}(t) \\ \omega_{r,w}(t) \end{bmatrix} = 
    U(p_{r,w}(t), \phi_{r,w}(t)) 
    \begin{bmatrix} \dot{p}_{r,w}(t) \\ \dot{\phi}_{r,w}(t) \end{bmatrix},
\end{equation}
where $U(p_{r,w}(t), \phi_{r,w}(t))$ is a matrix that depends on the choice of the parameterization. In particular, $U$ relates body velocity to the derivatives of the parameterization. We differentiate \eqref{eq:li_4.1-5} to yield
\begin{equation}
    \label{eq:li_4.1-6}
    \begin{bmatrix} \dot{v}_{r,w}(t) \\ \dot{\omega}_{r,w}(t) \end{bmatrix} = 
    U 
    \begin{bmatrix} \ddot{p}_{r,w}(t) \\ \ddot{\phi}_{r,w}(t) \end{bmatrix} +
    \dot{U} 
    \begin{bmatrix} \dot{p}_{r,w}(t) \\ \dot{\phi}_{r,w}(t) \end{bmatrix},
\end{equation}
which relates desired waypoints to body acceleration.
We define the position error as
\begin{equation}
    \begin{bmatrix} \widetilde{p} \\ \widetilde{\phi} \end{bmatrix} =
    \begin{bmatrix} p_{r,w} \\ \phi_{r,w} \end{bmatrix} - \begin{bmatrix} p^d_{r,w} \\ \phi^d_{r,w} \end{bmatrix}.
\end{equation}
To realize the desired body acceleration $(\ddot{p}^d_{r,w}, \ddot{\phi}^d_{r,w})$ throughout the motion, we define a control law based on feedback linearization, given by 
\begin{equation}
    \label{eq:li_4.1-8a}
    \begin{split}
    \tau &= J^\top H^\dagger \begin{bmatrix} M & 0 \\ 0 & I \end{bmatrix} \dot{U} \begin{bmatrix} \dot{p}_{r,w}(t) \\ \dot{\phi}_{r,w}(t) \end{bmatrix}\\
    &+ J^\top H^\dagger \begin{bmatrix} \omega_{r,w} \times Mv_{r,w} \\ \omega_{r,w} \times I\omega_{r,w} \end{bmatrix} \\
    &+ J^\top H^\dagger \begin{bmatrix} M & 0 \\ 0 & I \end{bmatrix} U \left(
    \begin{bmatrix} \ddot{p}^d_{r,w} \\ \ddot{\phi}^d_{r,w} \end{bmatrix} - K_D 
    \begin{bmatrix} \dot{\widetilde{p}} \\ \dot{\widetilde{\phi}} \end{bmatrix}
    - K_P \begin{bmatrix} \widetilde{p} \\ \widetilde{\phi} \end{bmatrix} 
    \right),
    \end{split}
\end{equation}
where $K_P$ and $K_D$ are proportional and derivative gain matrices. Both gain matrices must be positive definite.

We validate the performance of this controller by considering the error term dynamics. Applying the contact wrench term \eqref{eq:li_4-8} with the substitution \eqref{eq:li_4.1-6} to the relationship \eqref{eq:2-14} yields
\begin{equation}
    \label{eq:li_4.1-12}
    \begin{split}
    \tau &= J^\top(q)
    H^\dagger \biggl( 
    \begin{bmatrix} M & 0 \\ 0 & I \end{bmatrix} 
    U 
    \begin{bmatrix} \ddot{p}_{r,w}(t) \\ \ddot{\phi}_{r,w}(t) \end{bmatrix}\\
    &+ \dot{U} 
    \begin{bmatrix} \dot{p}_{r,w}(t) \\ \dot{\phi}_{r,w}(t) \end{bmatrix} +
    \begin{bmatrix} \omega_{r,w} \times Mv_{r,w} \\ \omega_{r,w} \times I \omega_{r,w} \end{bmatrix}
    \biggr).
    \end{split}
\end{equation}
By applying the control \eqref{eq:li_4.1-8a}, we obtain the closed-loop dynamic response
\begin{equation}
    \label{eq:li_4.1-14}
    \begin{split}
    H^\dagger \begin{bmatrix} M & 0 \\ 0 & I \end{bmatrix} U
    \begin{bmatrix} \ddot{p}^d_{r,w} \\ \ddot{\phi}^d_{r,w} \end{bmatrix} - 
    K_D 
    \begin{bmatrix} \dot{\widetilde{p}} \\ \dot{\widetilde{\phi}} \end{bmatrix}
    - K_P \begin{bmatrix} \widetilde{p} \\ \widetilde{\phi} \end{bmatrix}
    \\
    = H^\dagger     
    \begin{bmatrix} M & 0 \\ 0 & I \end{bmatrix} 
    U 
    \begin{bmatrix} \ddot{p}_{r,w}(t) \\ \ddot{\phi}_{r,w}(t) \end{bmatrix}.
    \end{split}
\end{equation}

Because $\begin{bmatrix} M & 0 \\ 0 & I \end{bmatrix}$ is positive definite and $U$ is nonsingular, \eqref{eq:li_4.1-14} simplifies to
\begin{equation}
    \label{eq:li_4.1-17}
    \begin{bmatrix} \ddot{\widetilde{p}} \\ \ddot{\widetilde{\phi}} \end{bmatrix} +
    K_D 
    \begin{bmatrix} \dot{\widetilde{p}} \\ \dot{\widetilde{\phi}} \end{bmatrix}
    + K_P \begin{bmatrix} \widetilde{p} \\ \widetilde{\phi} \end{bmatrix} = 0,
\end{equation}
which demonstrates that position error can be driven to zero with appropriate positive definite gain matrices.

\subsection{B. End-Effector Movement Control}
The desired position for a single end-effector at the end of boom $i$ is defined as the relative position of the $C_{bi}$ frame with respect to the $C_{si}$ frame. Tracking waypoints for the end-effector is accomplished using the standard PD controller
\begin{equation}
    \label{eq:PD_regrasp}
    \begin{bmatrix}
    f_{bi} \\ \tau_{\theta i}
    \end{bmatrix} = 
    K_P \begin{bmatrix} \widetilde{b}_i \\ 
    \widetilde{\theta}_i \end{bmatrix} + 
    K_D \begin{bmatrix} \dot{\widetilde{b}}_i \\ 
    \dot{\widetilde{\theta}}_i \end{bmatrix},
\end{equation}
where the pose error is defined as
\begin{equation}
    \begin{bmatrix} \widetilde{b}_i \\ 
    \widetilde{\theta}_i \end{bmatrix} = 
    \begin{bmatrix} b^d_i \\ \theta^d_i \end{bmatrix} -
    \begin{bmatrix} 
    b_i \\ \theta_i
    \end{bmatrix},
\end{equation}
and $f_{bi}$ and $\tau_{\theta i}$ defined in Figure~\ref{fig:boom_movement_model}. The gain matrices $K_P$ and $K_D$ in \eqref{eq:PD_regrasp} can be tuned separately for each stage of motion, so they are not necessarily the same as in \eqref{eq:li_4.1-8a}. 
This control strategy does not rely on a model of ReachBot, but in future work we will design controllers that exploit the model developed in Section~\ref{sec:model}C, for example model predictive controllers.

\section{Simulations} \label{sec:simulations}
In this section, we demonstrate ReachBot's mobility in a 2D world.
To showcase its unique ability to access challenging terrain, we deploy ReachBot in a simulated environment with sparse anchor points that would be impassible for existing robots.
ReachBot follows a series of waypoints under nominal conditions, as well under conditions that evoke common pitfalls of feedback linearization. To ensure the controller's robustness, we analyze ReachBot's performance while introducing modeling errors and process disturbances. By empirically demonstrating good performance under adverse conditions, we verify that our approach provides acceptable controllability.

\subsection{A. Simulation Setup}
The 2D simulation world is a hallway where the distance between the walls is less than the span of two of ReachBot's booms. Within the hallway, we define a discrete set of sparsely-spaced anchor points, with some adjacent anchor points up to several meters apart. The environment is defined this way such that a rigid-link articulated-arm robot would need to be prohibitively large to reach enough anchors to traverse the hallway. ReachBot's task is to move from one end of the hallway to the other.

\begin{figure}[tp!]
    \centering
   \subfloat[Body movement stage]{%
        \includegraphics[width=0.45\textwidth]{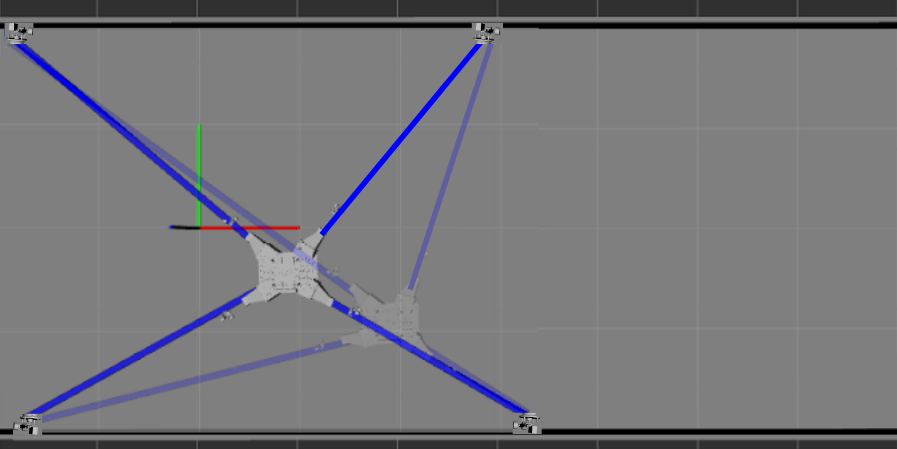}
        \label{fig:in_grasp}}
\hspace{\fill}
    \subfloat[End-effector movement stage]{%
      \includegraphics[width=0.45\textwidth]{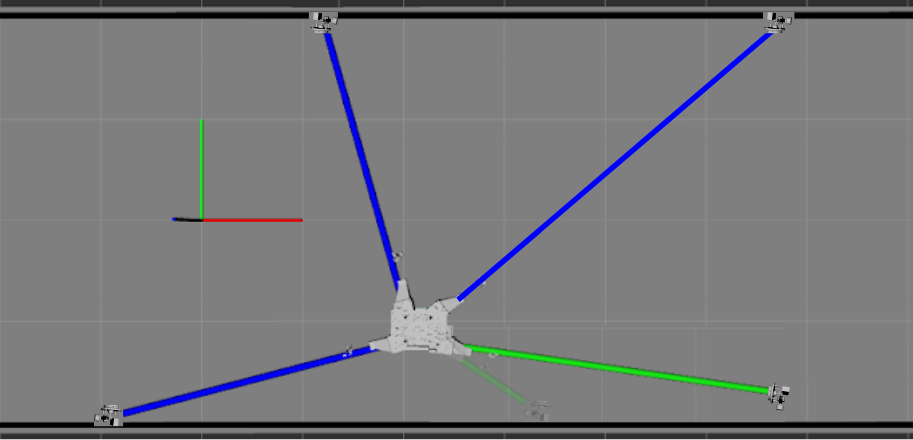}
      \label{fig:re_grasp}}
\caption{\label{fig:motion_stages} Snapshots of ReachBot simulation in different stages of motion. In the body movement stage (a), ReachBot's end-effectors remain anchored to the surface while the robot body moves in space. In the end-effector movement stage (b), ReachBot detaches an end-effector and moves it to a new anchor point. This mobility paradigm of alternating modes has been demonstrated successfully for articulated-arm robots~\protect\cite{Parness2017}.}
\vspace{-5pt}
\end{figure}

\begin{figure*}[htp!]
    \centering
        \includegraphics[width=0.95\textwidth]{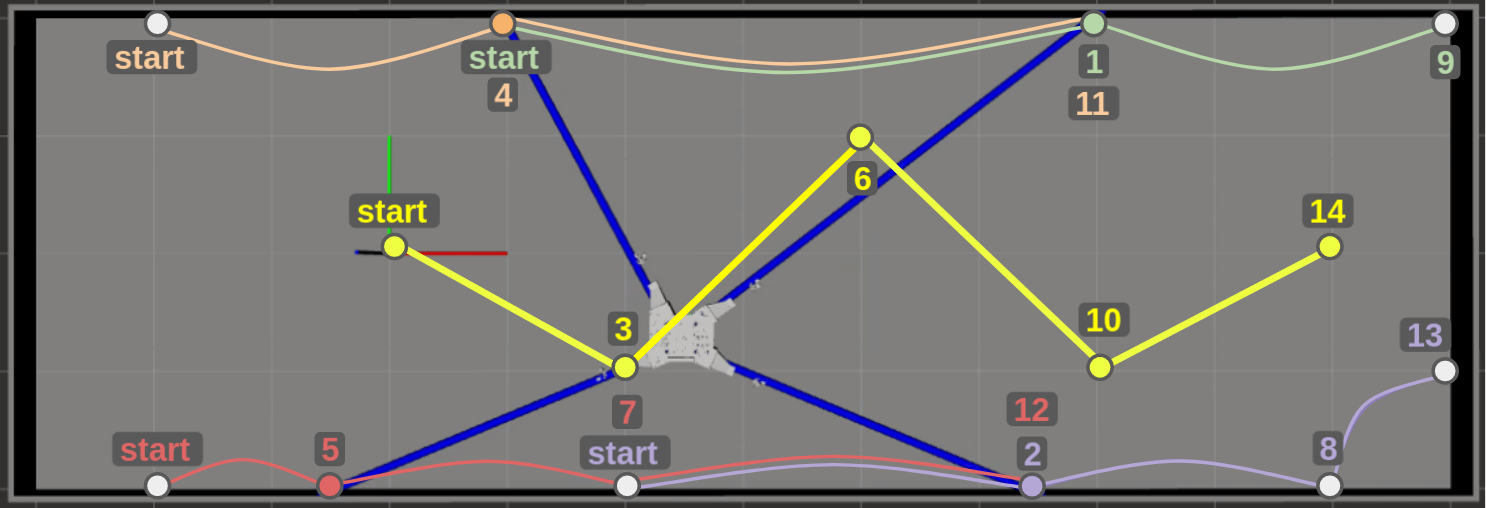}
        \caption{\label{fig:full_trajectory} Overview of ReachBot traversing a full sequence of waypoints. Desired waypoints and nominal trajectory are shown in yellow for the robot body and four separate colors for each of the four end-effectors. The sequence of waypoints is labeled in order, demonstrating the strategy of alternating body movement and end-effector movement to move from a starting state to a goal state.}
\end{figure*}

The simulation complements an ongoing hardware prototype, which is constrained by the size of the rollable booms. The following parameters are based on the specifications from that prototype.
ReachBot's body is a $30$cm x $20$cm x $20$cm rectangle with a mass of $30$kg and four extendable booms, attached to each corner by prismatic and planar revolute joints.
The actuator limits of these joints ($5$N for prismatic and $2.5$Nm for revolute) are enforced by the controller, which clips torque commands before sending them to the actuators.
Each end-effector has a mass of $1$kg. In simulation, the booms are assumed to be massless with a maximum extension of $5$m.
Additionally, in this planar environment we ignore gravity, as it would be constantly pushing into the plane. 

ReachBot's two alternating stages of motion are illustrated in Figure~\ref{fig:motion_stages}(a) for body movement and Figure~\ref{fig:motion_stages}(b) for end-effector movement.
Each waypoint defines a pose \textit{either} for the robot body \textit{or} a specific end-effector, therefore indicating the stage of motion at all times. The top-level planner uses a simple state machine to switch between the two controllers.
This approach allows ReachBot to track the full series of waypoints, and will serve as a baseline for trajectory optimization development in future work.

To avoid modeling complex contact interactions, 
we assume the end-effector waypoints correspond with secure anchor points.
When transitioning between stages, ReachBot must necessarily attach or detach an end-effector. To execute this transition, we assume the contacts remain within the friction limits to stay attached during body movement, and intentionally breach these limits to detach during end-effector movement. 
\begin{figure}[bp!]
    \centering
    \subfloat[Body position error over first minute of trajectory.]{%
      \includegraphics[width=0.95\columnwidth]{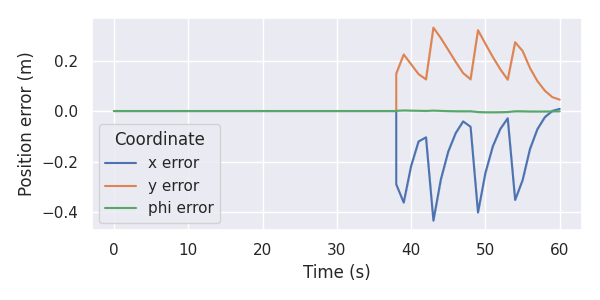}
      }
\hspace{\fill}
    \subfloat[End-effector position error over first minute of trajectory.]{%
      \includegraphics[width=0.95\columnwidth]{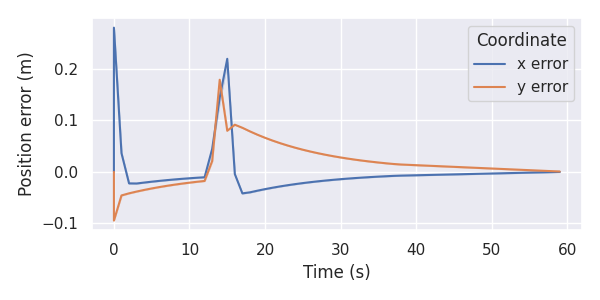}
      }
\caption{\label{fig:nominal} (a) The error between ReachBot's actual and nominal trajectory during body movement is quantified by the difference between the $x$, $y$, and $\phi$ position of the body. (b) During end-effector movement, the error between actual and nominal trajectory is given by the difference between $x$ and $y$ position of the moving end-effector.}
\end{figure}
\subsection{B. Results}
Figure~\ref{fig:full_trajectory} shows an overview of ReachBot's full sequence of waypoints, as well as the nominal trajectory between waypoints. The waypoints are shown in yellow for the robot body and four separate colors for each of the four end-effectors. The figure delineates the starting position of ReachBot and its four end-effectors, then labels each waypoint in order of the sequence. The full sequence demonstrates the strategy of alternating body movement and end-effector movement to reach a desired final goal state, in this case waypoint $14$.
Figures \ref{fig:nominal}(a) and (b) show position errors of the robot and active end-effector, respectively, during the first minute of the trajectory. This segment of the trajectory includes two end-effector movement and one body movement, up to waypoint $3$ in Figure~\ref{fig:full_trajectory}. The error plots spike as intermediate waypoints update instantaneously, which happens at a threshold distance of $5$mm and a threshold velocity of $2$mm/s. These error plots show ReachBot quickly converging to each waypoint.

\begin{figure}[bp!]
    \centering
     \includegraphics[width=0.95\columnwidth]{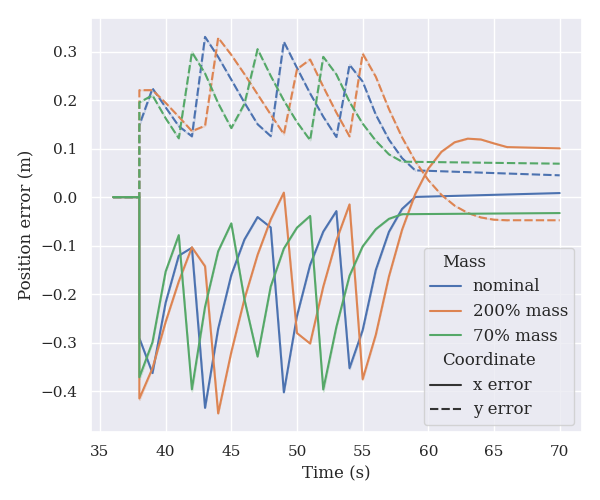}
    \caption{\label{fig:model_error} ReachBot body position error during the first body movement stage. 
    When the model underestimates mass (true mass is $200\%$ modeled mass magnitude), it takes longer for ReachBot to converge to the waypoint, in addition to suffering minor overshoot. However, both instances of model error reach the waypoint in reasonable time.}
\end{figure}

Controllers relying on feedback linearization are often susceptible to modeling errors, so we investigate this potential concern by injecting modeling errors and observing the system response.
Specifically, we adjust ReachBot's true mass and inertia without changing the modeled values.
Figure~\ref{fig:model_error} shows the system response for two scenarios: when the true mass is $70\%$ of the modeled mass, and when the true mass is $200\%$ of the modeled mass. In the former, the model overestimates ReachBot's mass, so the system expectantly reaches the waypoints faster than the nominal case. In the latter, underestimation of ReachBot's mass results in slower convergence and overshoot, but ReachBot still converges to the waypoints and follows the nominal trajectory.

\begin{figure}[hp!]
    \centering
    \includegraphics[width=0.95\columnwidth]{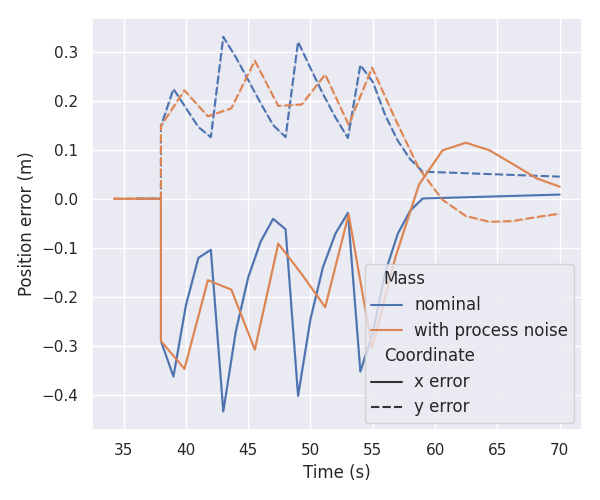}
    \caption{The plot of ReachBot's position error throughout body movement showcases the transients quickly dying out, even with the addition zero-mean Gaussian process noise.
    ReachBot still converges to each waypoint quickly, supporting the controller's robustness to process noise.}
    \label{fig:noise}
    \vspace{-5pt}
\end{figure}

ReachBot's controller must also be robust to process disturbances, so we similarly inject process noise and observe the system response. Figure~\ref{fig:noise} shows the robot's position error during the first body movement stage while injecting zero-mean Gaussian process noise with standard deviation of $\sigma = 0.05$, corresponding to about $5\%$ of the median control signal. During the course of the simulation, ReachBot deviates at most $20$cm from the nominal trajectory, averaging less than $5$cm deviation. Further, Figure~\ref{fig:noise} shows the deviations decaying quickly.
\subsection{C. Discussion}
Results from the simulation verify ReachBot's ability to follow a desired trajectory via a series of waypoints, even in the presence of modeling errors or process noise. 
With tuned gain matrices ($K_P$ and $K_D$ in \eqref{eq:li_4.1-8a} and \eqref{eq:PD_regrasp}), ReachBot converges quickly to the nominal trajectory,
minimizing the system's response time while avoiding both overshoot and overexerting the actuators.

\begin{figure}[tp]
    \centering
    \includegraphics[width=0.95\columnwidth]{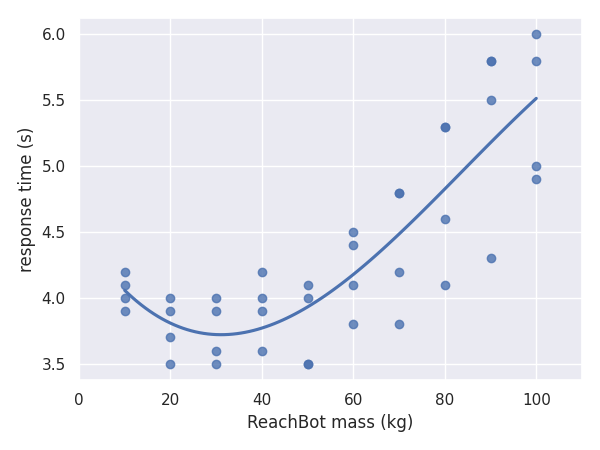}
    \caption{This figure highlights a fundamental trade-off between ReachBot mass and response time while holding actuator limits and other system specifications constant.
    The minimum response time is achieved with a weight of around $30$kg.}
    \label{fig:trade_study}
    \vspace{-5pt}
\end{figure}

Notably, the promising performance seen in Section~\ref{sec:simulations}B relies on our simulation parameters. For example, ReachBot's simulated mass ($30$kg) is on the same order as Axel or LEMUR designs for Martian exploration \textit{excluding} science instrument payloads, which can add up to $5$--$10$kg each~\cite{NesnasMatthewsEtAl2012,Parness2017}. Realistically, additional payloads, terrain features, and specific manipulation tasks could all place more demand on actuation.
Figure~\ref{fig:trade_study} displays response times for four different step movements while varying ReachBot's mass from $10$--$100$kg. Actuator joint limits ($5$N for prismatic and $2.5$Nm for revolute joints) begin inhibiting responsiveness at a mass of $60$kg, seen by the upward trend in response time.
The fastest response time is achieved with a weight of around $30$kg. Future work will include a full case study to consider system specification trade-offs, including boom configuration, system responsiveness, actuator strength, mass, and power requirements.
\vspace{-5pt}

\section{Conclusion}
In this paper, we propose a novel mobile manipulation robot that leverages extendable booms to navigate and explore its environment. These booms provide a lightweight, compact alternative to traditional robotic arms, enabling a robot with both increased reachable workspace and reduced complexity. ReachBot leverages these booms for mobility and manipulation tasks, fulfilling the technology gap for small mobile robots with high-wrench capabilities. In particular, ReachBot provides a solution to the mobility challenges of microgravity, very low gravity, or climbing under gravity, especially when environments have sparse anchor points.

ReachBot's novel use of extendable booms offers many advantages, including large reachable workspace and high-wrench capabilities. The use of long, lightweight booms presents structural challenges, which we address by leveraging tensile strength to support system stability.
We derive a dynamical model inspired by the grasp model for a dexterous manipulator, then modify an existing controller to enable ReachBot to track a desired series of waypoints. 
We demonstrate ReachBot's mobility with a pre-defined gait that alternates between ``body movement'' and ``end-effector movement'' stages. By validating our model in 2D simulation, we support ReachBot's ability to navigate terrain that is inaccessible to existing robots. 

Our initial results lay the foundation for extensive future work. First, 
the model and controller presented here will support experiments on a 2D prototype under concurrent development~\cite{ChenMillerEtAl2021}. Hardware experimentation will validate modeling assumptions and enable analysis for the structural capability of ReachBot's booms, even under different levels of gravity.
Depending on the mission and end-effectors, more specific models of contact are available~\cite{JiangWangEtAl2018} and will be incorporated in future work.
Additionally, we will use trajectory optimization and task and motion planning techniques to define trajectories that satisfy state, actuator, and structural constraints while maximizing system robustness.
Finally, we will perform a system-wide trade study to design a 3D configuration of ReachBot.
The results from this paper present a strong case for ReachBot's promising potential and provide a solid foundation for continued development.

\acknowledgments
Support for this work was provided by NASA under the Innovative Advanced Concepts program (NIAC).
Stephanie Schneider is supported by a NASA NSTGRO fellowship, and Tony G. Chen is supported by a NSF Graduate Fellowship. Thanks to Edward Schmerling and Abhishek Cauligi for their helpful comments.

\bibliographystyle{IEEEtran} 
\bibliography{ASL_papers,main,bib_standard_format}

\thebiography
\begin{biographywithpic}
{Stephanie Schneider}{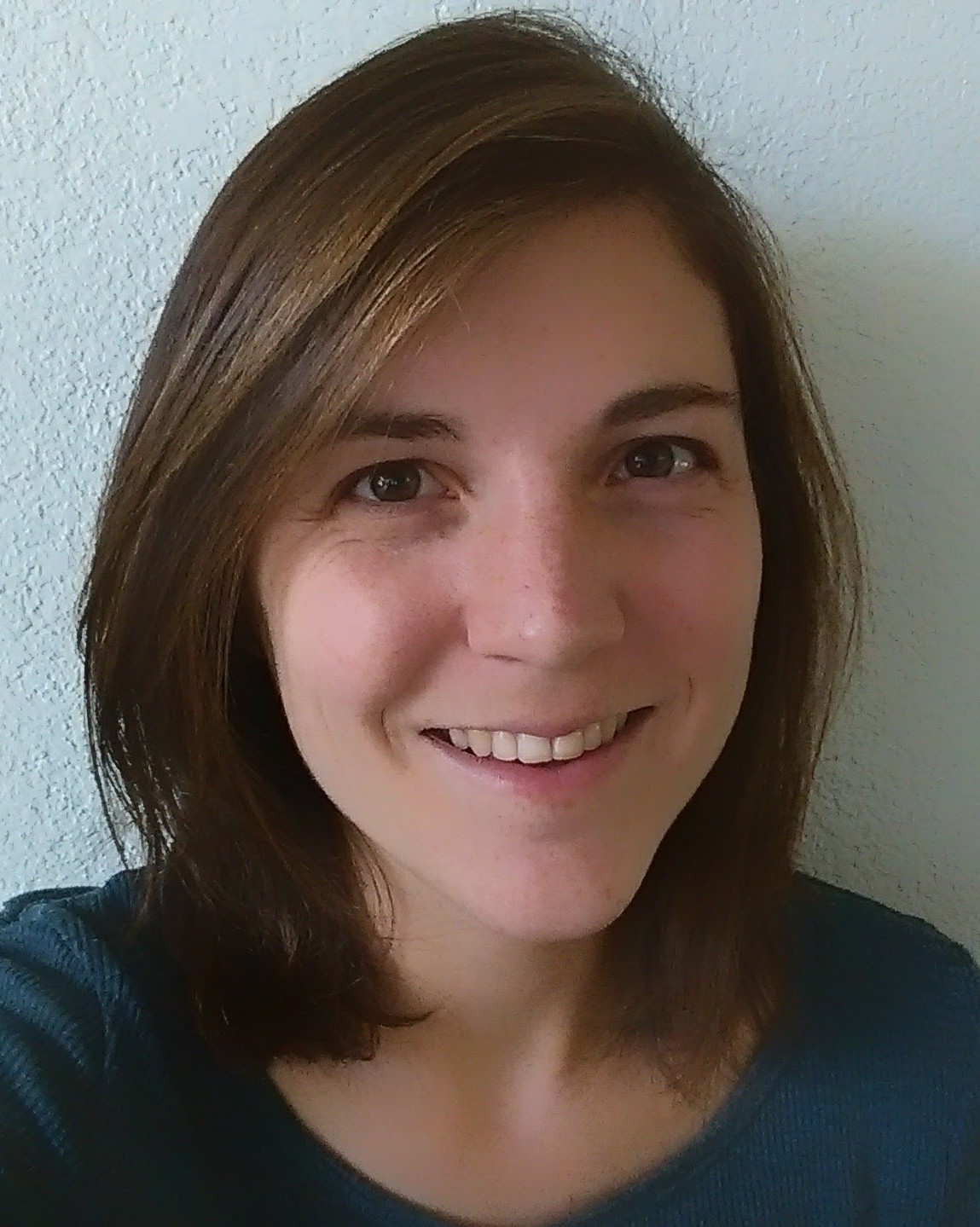}
is a Ph.D. candidate in the Autonomous Systems Lab in the Aeronautics and Astronautics Department at Stanford.
She received her B.S. in Mechanical Engineering from Cornell University in 2014. Prior to coming to Stanford, she worked as a software engineer and flight test engineer for Kitty Hawk Corporation (formerly Zee Aero, now Wisk). She is currently supported by an NSTGRO fellowship. Stephanie’s research interests include real-time spacecraft motion-planning, grasping and manipulation in space, and unconventional space robotics.
\end{biographywithpic}

\begin{biographywithpic}
{Andrew Bylard}{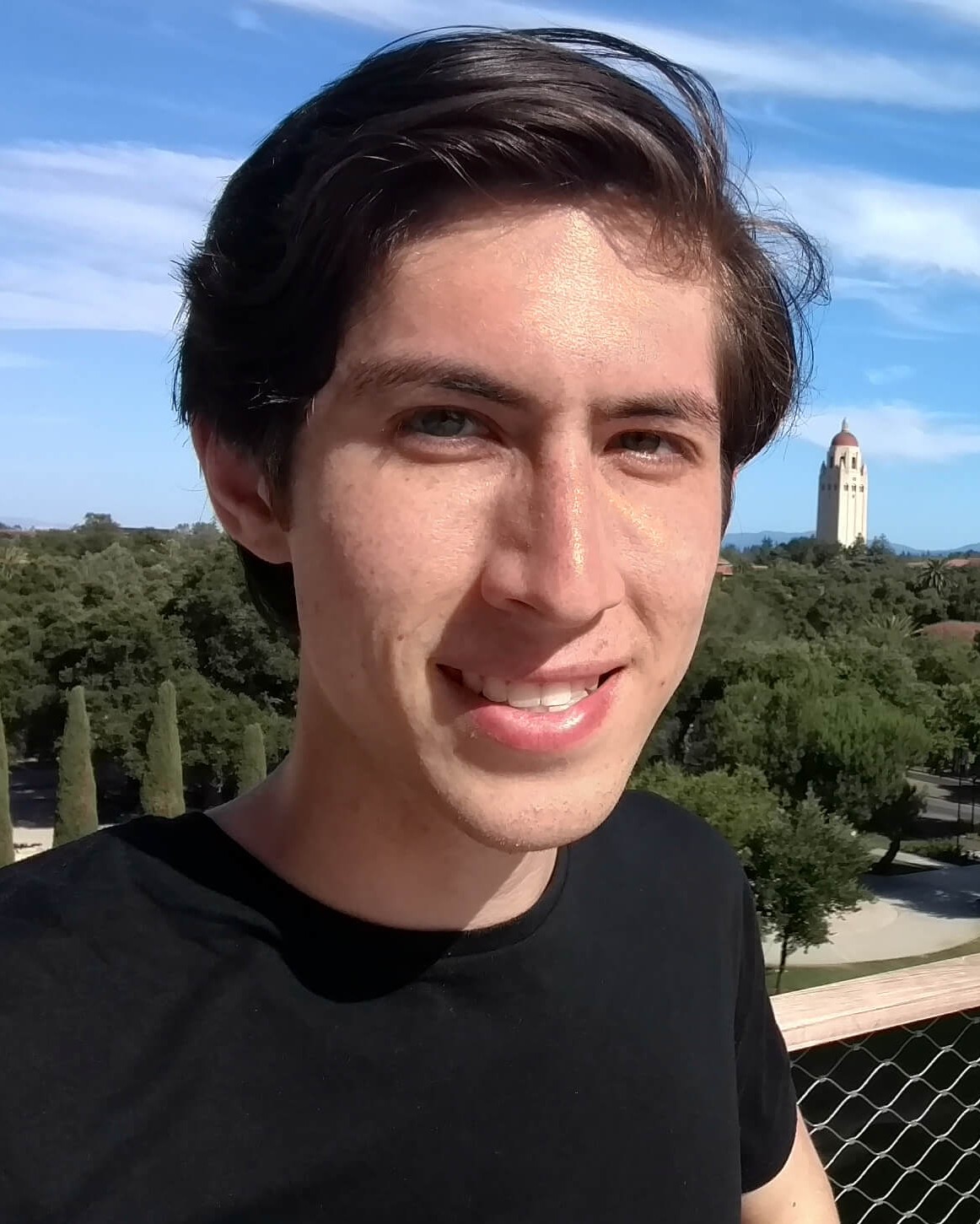}
is a Ph.D. candidate in Aeronautics and Astronautics at Stanford University. He received a B.Eng. with a dual concentration in Mechanical Engineering and Electrical Engineering from Walla Walla University in 2014 and a M.Sc. in Aeronautics and Astronautics from Stanford University in 2016. Andrew's research interests include real-time trajectory planning and optimization and unconventional space robotics, including using gecko-inspired adhesives for microgravity manipulation and repurposing rollable extendable booms for long-reach mobile manipulators in reduced gravity. He has been a lead developer at the Stanford Space Robotics Facility, where he designed robot test beds used to perform spacecraft contact dynamics experiments and develop autonomous space robotics capabilities under simulated frictionless, microgravity conditions.
\end{biographywithpic}

\begin{biographywithpic}
{Tony G. Chen}{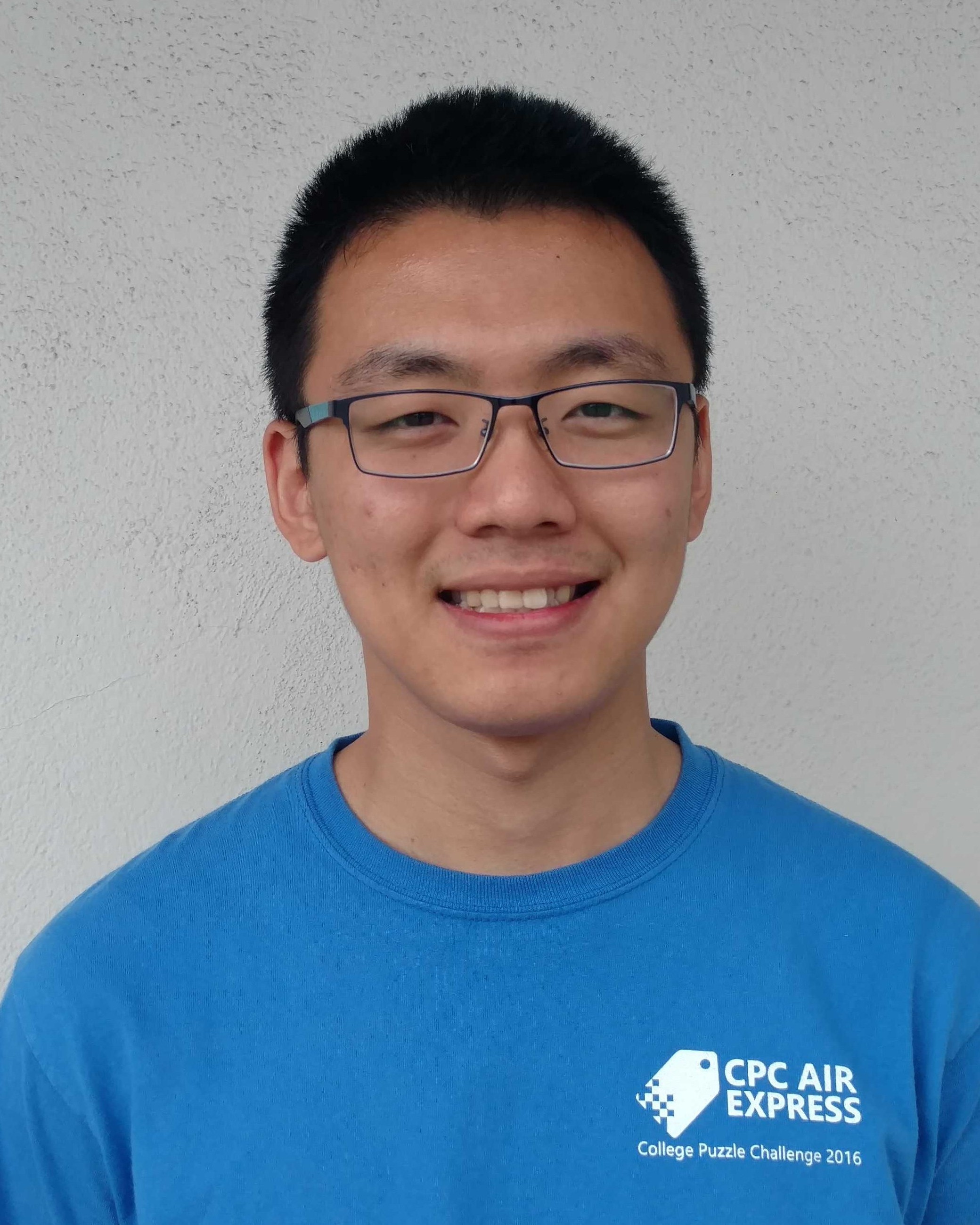}
is a Ph.D. candidate in the Biomimetics and Dexterous Manipulation Lab in the Mechanical Engineering Department at Stanford. He received his B.S. in Mechanical Engineering from Georgia Institute of Technology in 2017. Tony is a NASA Robotics Academy graduate and interned at NASA JPL in the Extreme Environment Robotics group working on the Asteroid Redirect Mission. Tony's research interests include bio-inspired and field robotics, particularly designing climbing and perching robots that interact with challenging, real-world environments. 
\end{biographywithpic}

\begin{biographywithpic}
{Preston Wang}{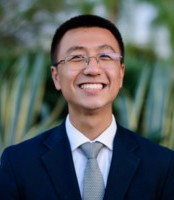}
graduated with a M.Sc. degree from the Aeronautics and Astronautics Department at Stanford, where he worked on structural analysis of a ReachBot robot in climbing applications and developed an initial 3-meter extending and pivoting ReachBot arm prototype that could grasp and manipulate loads under earth gravity. He now works as a propulsion engineer at SpaceX.
\\
\\
\\
\end{biographywithpic}

\begin{biographywithpic}
{Mark Cutkosky}{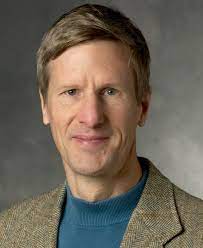}
is the director of the Biomimetics and Dexterous Manipulation Lab in the Mechanical Engineering Department at Stanford. He has been active in bio-inspired robots, dexterous manipulation and rapid prototyping since 1985. His research on insect- and gecko-inspired adhesives resulted in the Spinybot and Stickybot climbing robots, and has been applied to human climbing, perching micro air vehicles and micro robots. Cutkosky is a fellow of IEEE and ASME, a former National Science Foundation Presidential Young Investigator and a former Fulbright Distinguished Chair.
\end{biographywithpic}

\begin{biographywithpic}
{Marco Pavone}{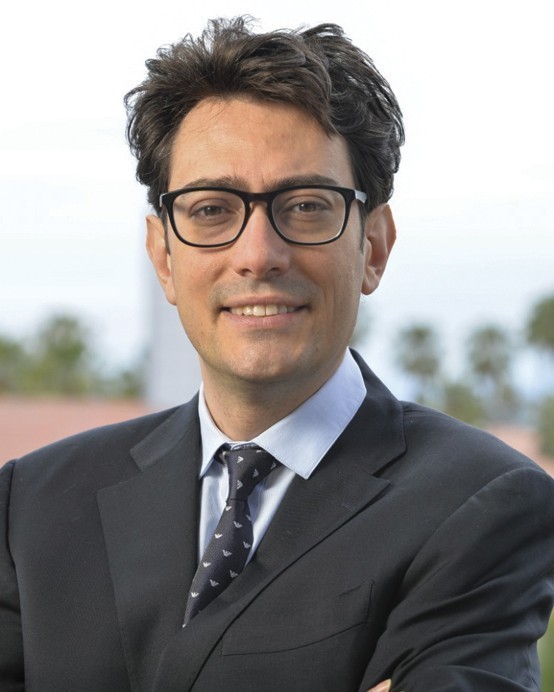}
is an Associate Professor of Aeronautics and Astronautics at Stanford
University, where he is the Director of the Autonomous Systems Laboratory. Before joining Stanford, he was a Research Technologist within the Robotics Section at the NASA Jet Propulsion Laboratory. He received a Ph.D. degree in Aeronautics and Astronautics from the
Massachusetts Institute of Technology in 2010. Dr. Pavone’s expertise lies in the fields of controls and robotics. His main research interests are in the development of methodologies for the analysis, design, and control of autonomous systems, with an emphasis on autonomous aerospace vehicles and large-scale robotic networks.
\end{biographywithpic}

\end{document}